\pdfoutput=1

\documentclass[11pt]{article}

\usepackage{ACL2023}

\usepackage{times}
\usepackage{latexsym}

\usepackage[T1]{fontenc}

\usepackage[utf8]{inputenc}

\usepackage{microtype}

\usepackage{inconsolata}

\usepackage{graphicx}
\usepackage{microtype}
\usepackage{enumitem}
\usepackage{algorithm}
\usepackage{algorithmic}
\usepackage{amsmath}
\usepackage{multirow}
\usepackage{booktabs}
\usepackage{url}
\usepackage{caption}
\usepackage{subcaption}
\usepackage{listings}
\usepackage{tablefootnote}
\usepackage{adjustbox}

\definecolor{codegreen}{rgb}{0,0.6,0}
\definecolor{codegray}{rgb}{0.5,0.5,0.5}
\definecolor{codepurple}{rgb}{0.58,0,0.82}
\definecolor{backcolour}{rgb}{0.95,0.95,0.92}

\lstdefinestyle{mystyle}{
    backgroundcolor=\color{backcolour},   
    commentstyle=\color{codegreen},
    keywordstyle=\color{magenta},
    numberstyle=\tiny\color{codegray},
    stringstyle=\color{codepurple},
    basicstyle=\ttfamily\footnotesize,
    breakatwhitespace=false,         
    breaklines=true,                 
    captionpos=b,                    
    keepspaces=true,                 
    numbers=left,                    
    numbersep=5pt,                  
    showspaces=false,                
    showstringspaces=false,
    showtabs=false,                  
    tabsize=2
}

\usepackage{xcolor}
\usepackage{color, colortbl}
\definecolor{light-gray}{gray}{0.95}
\definecolor{antiquewhite}{rgb}{0.98, 0.92, 0.84}
 	
\usepackage{soul}

\newcommand{\hlc}[2][yellow]{{%
    \colorlet{foo}{#1}%
    \sethlcolor{foo}\hl{#2}}%
} 	
 	
\newcommand{\code}[1]{\colorbox{light-gray}{\texttt{#1}}}

\NewDocumentCommand{\alex}{ mO{} }{\textcolor{red}{\textsuperscript{\textit{Alex}}\textsf{\textbf{\small[#1]}}}}

\NewDocumentCommand{\eva}{ mO{} }{\textcolor{blue}{\textsuperscript{\textit{Eva}}\textsf{\textbf{\small[#1]}}}}

\NewDocumentCommand{\jie}{ mO{} }{\textcolor{green}{\textsuperscript{\textit{Jie}}\textsf{\textbf{\small[#1]}}}}

\NewDocumentCommand{\mingyue}{ mO{} }{\textcolor{orange}{\textsuperscript{\textit{Mingyue}}\textsf{\textbf{\small[#1]}}}}

\NewDocumentCommand{\patricng}{ mO{} }{\textcolor{magenta}{\textsuperscript{\textit{Patrick}}\textsf{\textbf{\small[#1]}}}}

\NewDocumentCommand{\zhiguow}{ mO{} }{\textcolor{purple}{\textsuperscript{\textit{Zhiguo}}\textsf{\textbf{\small[#1]}}}}

\NewDocumentCommand{\vittorio}{ mO{} }{\textcolor{brown}{\textsuperscript{\textit{Vittorio}}\textsf{\textbf{\small[#1]}}}}

\title{Few-Shot Data-to-Text Generation via Unified Representation and Multi-Source Learning}

\author{Alexander Hanbo Li, Mingyue Shang, Evangelia Spiliopoulou, Jie Ma \\
        {\bf Patrick Ng, Zhiguo Wang, Bonan Min, William Wang} \\
        {\bf Kathleen McKeown, Vittorio Castelli, Dan Roth, Bing Xiang} \\
        AWS AI Labs \\
        \texttt{\footnotesize\{hanboli, myshang, spilieva, jieman, patricng, zhiguow, bonanmin, wyw\}@amazon.com} \\
        \texttt{\footnotesize\{mckeownk, vittorca, drot, bxiang\}@amazon.com}}

\begin{document}
\maketitle
\begin{abstract}

We present a novel approach for structured data-to-text generation that addresses the limitations of existing methods that primarily focus on specific types of structured data. Our proposed method aims to improve performance in multi-task training, zero-shot and few-shot scenarios by providing a unified representation that can handle various forms of structured data such as tables, knowledge graph triples, and meaning representations. We demonstrate that our proposed approach can effectively adapt to new structured forms, and can improve performance in comparison to current methods. For example, our method resulted in a 66\% improvement in zero-shot BLEU scores when transferring models trained on table inputs to a knowledge graph dataset. Our proposed method is an important step towards a more general data-to-text generation framework. 
\end{abstract}

\section{Introduction}
\label{sec:intro}

Data-to-text generation is the task of converting structured data into natural language text that can be easily understood by humans. Previous methods for data-to-text generation have been limited to specific structured forms. For example, graph neural networks (GNNs) have been used to encode knowledge graph input \citep{koncel2019text,kg2text-2020,guo-etal-2020-cyclegt,li-etal-2021-shot-knowledge}, while table-specific encoders have been proposed for tables \citep{liu2017table,Bao2018TabletoTextDT,nema-etal-2018-generating,jain-etal-2018-mixed,wang-etal-2022-robust}. However, these methods are not easily transferable to other structured forms, creating a barrier for scientific development and preventing models from learning across tasks. Recent work has attempted to address the problem of limited structured form applicability by using pretrained language models (PLMs) as a single text-to-text framework for all data structures, by linearizing the data as text sequences. As shown by \citet{kale-rastogi-2020-text,UnifiedSKG}, these methods achieve state-of-the-art performance on a wide range of data-to-text tasks. 

Despite the advancements made in the field, there are still unresolved questions regarding the relationship between various structured forms, particularly in the context of zero-shot or few-shot settings, where models are required to rapidly adapt to new structured forms. This is particularly pertinent in cases of data scarcity, when structured forms vary across different domains and there is a limited amount of data available for a specific structured form, but a single model is needed to operate on all of them. Such an example is to adapt a knowledge-graph-to-text model to a new domain with data in table format. Even when there is an abundance of data, developing a universal model that can handle all structured forms remains a challenging task. As seen in \citet{UnifiedSKG}, a multi-task trained model may perform worse than a single-task model on table inputs. One important reason for such performance drop is because previous research has not fully examined the impact of various linearization methods on these tasks and their effect on cross-task generalization. Despite the use of text-to-text transformers, linearization methods for various structured forms remain diverse, and even within one structured form, linearization can vary across studies. For example, the linearization of KG triples differs in \citet{nan-etal-2021-dart} and \citet{UnifiedSKG}, highlighting the need for further research on the relationship between data formats and data-to-text tasks.

In this paper, we address the unresolved questions surrounding the relationship between various structured forms by introducing a \textit{unified representation} for knowledge graphs, tables, and meaning representations. We demonstrate that our method allows for the conversion of knowledge graph triples and meaning representations into virtual tables, which can then be linearized in a consistent manner. Through evaluating our approach on five representative data-to-text tasks across the aforementioned formats, we show that our method not only achieves competitive performance compared to other data-specific linearizations for individual tasks, but also leads to significant improvements in transfer learning scenarios across structured forms, particularly in zero-shot or few-shot settings. For example, using the unified representation improves the zero-shot BLEU score by relatively 66\% when transferring from ToTTo \citep{parikh2020totto} to DART \citep{nan-etal-2021-dart}. Additionally, our approach results in improved performance when used in multi-task settings compared to models trained with varied linearizations. These results provide a clear indication of the effectiveness of our proposed unified representation in enhancing cross-task generalization.
\section{Related Work}
\label{sec:related_work}


\paragraph{Data-Type Specific Knowledge Encoding}

Research has been conducted to encode structured knowledge using various models and approaches, including Graph Neural Networks (GNNs) \citep{koncel2019text,kg2text-2020,guo-etal-2020-cyclegt,li-etal-2021-shot-knowledge,song-etal-2018-graph,ribeiro-etal-2019-enhancing,Cai_Lam_2020,zhang-etal-2020-lightweight,amr-to-text-2021,schmitt-etal-2021-modeling} and neural encoder-decoder models based on Gated Recurrent Units (GRUs) and Transformers \citep{gehrmann2018end,ferreira2019neural}. These models have been used to assist in encoding knowledge graph inputs and meaning representations. Additionally, several models have been proposed for table-to-text generation, including approaches that combine content selection or entity memory in a Long Short-Term Memory (LSTM) model \citep{puduppully2018data,puduppully2019data}, and others that focus on table-specific encoders \citep{liu2017table,Bao2018TabletoTextDT,nema-etal-2018-generating,jain-etal-2018-mixed}. More recent studies have utilized the capabilities of pre-trained language models in their designs, but have also incorporated specialized encoder structures or attention mechanisms specifically for table inputs. These include encoder-only models \citep{Arik2019TabNetAI,yin-etal-2020-tabert,herzig-etal-2020-tapas,Huang2020TabTransformerTD,tuta-2021,iida-etal-2021-tabbie,eisenschlos-etal-2021-mate,tableformer}, as well as encoder-decoder models \citep{Cao2020GeneratingNL,tabt5-2022,wang-etal-2022-robust}. However, it should be noted that the encoder structures of these works are specifically tailored for table input and cannot be directly applied to other types of data.

\paragraph{Structured Data Linearization}
Recent developments in pretrained language models \citep{devlin-etal-2019-bert,Radford2019LanguageMA,lewis-etal-2020-bart,2020t5} have made it possible to use a single text-to-text framework for various types of data by linearizing them as text sequences. Studies have been conducted on finetuning PLMs on table input \citep{parikh2020totto} and knowledge graph input \citep{kasner-dusek-2020-train,ribeiro-etal-2021-investigating}, single-task and multi-task training on a collection of structured data grounding tasks \citep{UnifiedSKG}, and the effectiveness of pretraining and fine-tuning strategies for data-to-text tasks \citep{kale-rastogi-2020-text} and table-based question answering tasks \citep{Shi2022GenerationFocusedTI}. These studies have consistently found that linearizing structured data as a sequence of tokens without modifying the model structure, is a simple yet effective strategy that outperforms pipelined neural architectures specifically tailored to particular data types.





\paragraph{Zero/Few-Shot Data-to-Text Generation}
The studies such as \citet{chen-etal-2020-kgpt} and \citet{ke-etal-2021-jointgt} have evaluated the zero and few-shot performance of PLMs on knowledge graph input, highlighting the benefits of a joint pretraining strategy on knowledge graphs and texts for learning better KG representations. \citet{Keymanesh2022WhatMD} studied the prompt-tuning method for KG-to-text generation and found it to be effective in a few-shot setting. \citet{chen-etal-2020-shot} combines PLM with a table content selector using a switch policy. Other researchers have also explored methods such as data augmentation \citep{chang-etal-2021-neural} and retrieval-based input augmentation \citep{su-etal-2021-shot-table} to aid in few-shot data-to-text generation. \citet{kasner-dusek-2022-neural} proposes a pipeline approach involving a sequence of operations, such as ordering and aggregation, and only finetunes the PLMs of these modules to make the pipeline more domain-independent. 
\section{Unified Representation}
\label{sec:model}

\begin{figure*}[hbt]
    \centering
    \includegraphics[width=0.90\textwidth]{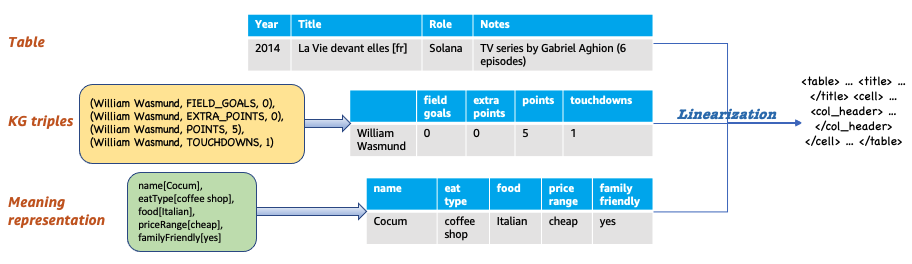}
    \caption{unified representation of three data types: table, KG triples, and meaning representations. The latter two are first converted to virtual tables, and then linearized using the same method as table input.}
    \label{fig:method}
\end{figure*}

In this section, we demonstrate that structured data, such as tables, highlighted cells, knowledge graph triples, and meaning representations, can be linearized in a consistent manner. We begin by showing in Section \ref{ssec:virtual-table} how knowledge graph triples and meaning representations can be mapped to a virtual table and subsequently linearized in the same way as tables. Next, in Section \ref{ssec:linearize-table}, we demonstrate the process of linearizing a table or highlighted cells. The entire method is illustrated in Figure \ref{fig:method}.

\subsection{Virtual Table}
\label{ssec:virtual-table}

\paragraph{KG Triple}
\label{sssec:linearize-kg}
The method for converting triples from a connected sub-graph into a virtual table involves using the tail node of each triple as a cell value and the relation as the column header. Nodes that do not appear as tail nodes are not assigned a column header. An example is provided in Figure \ref{fig:method}. "William Wasmund" does not have a column header assigned since it never appears as a tail node. If a set of knowledge graph triples contains multiple connected components, each component is converted into a separate table.



\paragraph{Meaning Representation}
\label{sssec:linear-mr}
We focus on textual MRs that appear as a list of comma-separated attribute-value pairs \citep{e2e-2020}. These MRs can be treated as virtual tables by associating each \texttt{Attribute[Value]} with a cell value, represented by the "Value", and the "Attribute" as its corresponding column header. An example of this can be seen in Figure \ref{fig:method}. 



\subsection{Linearization of Tables}
\label{ssec:linearize-table}
After converting both KGs and MRs into virtual tables, we end up with only table inputs that need to be linearized. In this section, we discuss one choice of such a linearization method, motivated by ToTTo linearization \citep{parikh2020totto}. Additionally, we will provide a specific example of how to linearize Table \ref{tab:example} in the following sections.

\begin{table}[htb]
\centering
\footnotesize
\begin{tabular}{p{0.10\linewidth} | p{0.47\linewidth} | p{0.15\linewidth}}
\multicolumn{3}{l}{\textbf{Table Title}: Alma Jodorowsky}   \\
\multicolumn{3}{l}{\textbf{Section Title}: Filmography} \\ \toprule
\textbf{Year}   & \textbf{Title}  & \textbf{Role}     \\ 
\midrule
2014   & La Vie devant elles [fr]   & Solana   \\ 
\hline
\cellcolor[HTML]{FFEF00}2016   & \cellcolor[HTML]{FFEF00}Kids in Love   & \cellcolor[HTML]{FFEF00}Evelyn    \\ 
\hline
2017   & The Starry Sky Above Me   & Justyna    \\ 
\bottomrule
\end{tabular}
\caption{An example table to showcase our linearization.}
\label{tab:example}
\end{table}

\paragraph{Basic Units}
The basic units for linearization are presented in Table \ref{tab:basic-units}. Each unit is defined by a start symbol, \texttt{<xx>}, and an end symbol, \texttt{</xx>}.

\begin{table}[htb]
\centering
\footnotesize
\begin{tabular}{p{0.28\linewidth} | p{0.60\linewidth}}
\toprule
\textbf{Start Symbol} & \textbf{Meaning} \\
\midrule
\code{<table>} & contents in a table \\

\code{<column>} &  contents in a column \\

\code{<row>} & contents in a row \\

\code{<cell>} & content in a cell \\

\code{<col\_header>} & column header name \\

\code{<row\_header>} & row header name \\

\code{<title>} & main title /domain / topic of the input \\

\code{<sub\_title>} & sub-title /domain /topic of the input \\
\bottomrule
\end{tabular}
\caption{Basic units of our linearization.}
\label{tab:basic-units}
\end{table}

\paragraph{Linearization of Highlighted Cells}
\label{sssec:linearize-highlighted}

To linearize the highlighted cells, we proceed in a left-to-right, top-to-bottom order. For instance, in Table \ref{tab:example}, the linearization of the highlighted cells (in yellow background) appears as follows: \footnote{Indentation is used for clarity in this example, but it is not present in the actual input.}

\begin{lstlisting}[language=Html, moredelim={[is][\color{magenta}]{START}{END}}]
<title> Alma Jodorowsky </title>
<STARTsub_titleEND> Filmography <START/sub_titleEND>
<table> 
  <STARTcellEND> 2016 
    <STARTcol_headerEND> Year <START/col_headerEND>
  <START/cellEND>
  <STARTcellEND> Kids in Love 
    <STARTcol_headerEND> Title <START/col_headerEND>
  <START/cellEND>
  <STARTcellEND> Evelyn
    <STARTcol_headerEND> Role <START/col_headerEND>
  <START/cellEND>
 </table>
\end{lstlisting}

\paragraph{Linearization of (Sub)Table}
\label{sssec:linearize-subtable}
A row-wise linearization of the entire Table \ref{tab:example} is:

\adjustbox{max width=\textwidth}{%
\begin{lstlisting}[language=Html, moredelim={[is][\color{magenta}]{START}{END}}]
<title> Alma Jodorowsky </title>
<STARTsub\_titleEND> Filmography <START/sub\_titleEND>
<table>
  <STARTrowEND>
    <STARTcellEND> 2014
      <STARTcol_headerEND> Year <START/col_headerEND>
    <START/cellEND>
    <STARTcellEND> La Vie devant elles [fr]
      <STARTcol_headerEND> Title <START/col_headerEND>
    <START/cellEND>
    <STARTcellEND> Solana
      <STARTcol_headerEND> Role <START/col_headerEND>
    <START/cellEND>
  <START/rowEND>
  ...(other rows)... 
</table>
\end{lstlisting}
}

Such a linearization method can also be applied to column-wise. An example is provided in the Appendix~\ref{col-wise-example}.



\section{Experiments}
\label{sec:experiments}

\paragraph{Datasets}
We test our method on five data-to-text datasets:
The \textbf{ToTTo} dataset \citep{parikh2020totto} poses the challenge of generating a one-sentence description, given highlighted cells from a Wikipedia table. Our models are evaluated on the validation set, as the annotations for the test set are not publicly available.
The \textbf{DART} corpus \citep{nan-etal-2021-dart} is an open-domain structured data-to-text resource, consisting of entity-relation triples.
The \textbf{LogicNLG} dataset \citep{chen-etal-2020-logical} investigates the ability to generate logical inferences from table contents to implicit insights, as the target sentences.
The \textbf{WebNLG} dataset \citep{gardent-etal-2017-webnlg} includes triples from 15 DBpedia categories, which are mapped to their verbalization. Results are reported on the Seen (S), Unseen (U), and All (A) subsets of the data.
The \textbf{E2E clean} dataset \citep{dusek-etal-2019-semantic} consists of meaning representations (MRs) from the restaurant domain. The task is to generate a sentence that verbalizes the \textit{useful} information from the MR.
Dataset statistics are summarized in Table \ref{tab:data} in the appendix.

\paragraph{Evaluation Metrics}
We evaluate the quality of generated texts using several widely accepted metrics. \textit{BLEU} \citep{papineni-etal-2002-bleu} measures the similarity between generated text and references in terms of n-gram overlap. \textit{METEOR} \citep{banerjee-lavie-2005-meteor} assesses the quality of generated text by comparing unigram matches between the text and references, including exact, stem, synonym, and paraphrase matches. \textit{TER} \citep{snover-etal-2006-study} is a measure of the number of edits required to change the generated text into one of the references. \textit{PARENT} \citep{parent-metric-2019} takes into account the table input when evaluating generated text. \textit{NIST} \citep{nist-2002} is similar to BLEU, but also considers the informativeness of each n-gram. \textit{CIDEr} \citep{vedantam2015cider} uses TF-IDF to lower the weights of common n-grams that appear in all references when calculating uni-gram to 4-gram overlaps between generated and reference sentences. We also use the \textit{NLI score} \citep{chen-etal-2020-logical} on the LogicNLG dataset to evaluate the logical fidelity, which is a model-based evaluation using the BERT model trained on the TabFact \citep{Chen2020TabFact} dataset.
    
\paragraph{Comparing Linearizations} 
\label{ssec:linear}
We compare our proposed \textit{unified representation} to other linearization methods from previous papers. Specifically, on DART, WebNLG, and E2E datasets, we compare our method to the linearization used in UnifiedSKG \citep{UnifiedSKG}.\footnote{The E2E dataset is not studied in the paper, but the linearization is included in their \href{https://github.com/HKUNLP/UnifiedSKG/blob/main/seq2seq_construction/e2e_nlg_cleaned.py}{official repository.}} On ToTTo and LogicNLG datasets, we use the linearization from their original papers \citep{parikh2020totto,chen-etal-2020-logical} for comparison. Examples of their linearization methods can be found in the appendix.

\subsection{Zero and Few-Shot Experiments}
Our hypothesis is that a model trained on one structured form will transfer better to other forms under zero or few-shot settings when using our unified method of representation. We test this by focusing on transferring from ToTTo data (table input) to other types and from WebNLG (KGs) to ToTTo in this section. Results for other transfers can be found in the appendix.
\label{ssec:task-transfer}
\begin{table}[htb]
\centering
\footnotesize
\scalebox{0.9}{
\begin{tabular}{lcc}
\toprule
\textbf{Setting} & \textbf{Src representation} & \textbf{Tgt representation} \\
\midrule
\textit{Only on tgt}         & -       & Others \\
\textit{Src to tgt, unified} & Unified & Unified \\
\textit{Src to tgt, varied}  & Others   & Others \\
\bottomrule
\end{tabular}}
\caption{Comparison of source and target task representations. "Unified" uses our proposed unified representation, "Others" uses linearizations from other papers for each task.}
\label{tab:settings}
\end{table}

As shown in Table \ref{tab:settings}, for each experiment, we compare \textbf{three settings}: 
(i) \textit{Only on tgt} -- In few-shot experiments, we only train the model on the target task using the linearization from other papers. In zero-shot experiments, we use the foundational model without any training.
(ii) \textit{Src to tgt, unified} -- First, train the model on the source task and then fine-tune it on $k$-shot\footnote{$k=0$ means no training on target task at all.} target-task data, using our unified representation for both.
(iii) \textit{Src to tgt, varied} -- Similar to (ii), but we use the linearization from other papers for each task, as described in \ref{ssec:linear}. We refer to this as the varied setting because the source and target-task linearizations are different.

During inference, we apply the same linearization method utilized during training to each target task. More implementation details are presented in the appendix.

\begin{table*}[hbt]
\centering
\footnotesize
\scalebox{0.9}{
\begin{tabular}{l|ccc|ccc|ccc|cc}
\toprule
    \multirow{2}{*}{\textbf{Setting}} & \multicolumn{3}{c}{ \textbf{DART (KG)}}  &  \multicolumn{3}{c}{ \textbf{WebNLG (KG)}} & \multicolumn{3}{c}{ \textbf{E2E clean (MR)}} & \multicolumn{2}{c}{ \textbf{LogicNLG (Table)}} \\
    &  \textbf{BLEU} & \textbf{MET} &\textbf{TER} $\downarrow$ & \textbf{S} & \textbf{U} &\textbf{A} & \textbf{BLEU} & \textbf{NIST} &\textbf{CIDEr} & \textbf{BLEU-3} & \textbf{NLI} \\
\midrule
    GPT2-XL  & 13.3 & 0.24 & 0.65 & - & - & - & - & - & - & - & - \\
    KGPT & - & - & - & - & - & 13.9 & - & - & - & - & - \\
    JointGT (0.5\%)$^a$ & - & - & - & - & 37.2 & - & - & - & - & - & - \\
    HTLM (1-shot)$^a$ & 22.1 & 0.12 & 0.91 & 28.1 & 18.5 & 22.8 & - & - & - & - & - \\
\midrule
    Only on tgt$^b$ & 0.3    & 0.01  & 2.82 & 0.36 & 0.08 & 0.23 & 0.0 & 0.0 & 0.0 & 0.2 & \textbf{85.1} \\
    Src to tgt, varied         & 18.9   & 0.21  & 1.00 & 34.1 & 28.5 & 31.3 & 12.1 & 2.8 & 0.3 & 7.8 & 70.9 \\
    Src to tgt, unified  & \textbf{31.5}   & \textbf{0.32}  & \textbf{0.56} & \textbf{35.9} & \textbf{39.8} & \textbf{37.7} & \textbf{22.6} & \textbf{4.4} & \textbf{0.9} & \textbf{8.9} & 81.3\\
\bottomrule
\multicolumn{12}{l}{$^a$ \textit{We compare our results to their few-shot performance, as zero-shot results are not reported in their papers.}}\\
\multicolumn{12}{l}{$^b$ \textit{Under zero-shot, this means directly testing T5-base model on target test set without any training.}}\\
\end{tabular}}
\caption{Zero-shot results. Our foundational model is T5-base (220M). MET stands for METEOR, and lower scores on TER indicate better performance. On WebNLG, BLEU scores are reported for seen (S), unseen (U), and all (A) categories. The NLI-accuracy is calculated using the NLI model provided in LogicNLG official codebase. On papers without zero-shot results, we report their few-shot performance.}
\label{tab:zero-shot}
\end{table*}

\begin{figure*}[!htb]
     \centering
     \begin{subfigure}{0.31\textwidth}
         \centering
         \includegraphics[width=\textwidth]{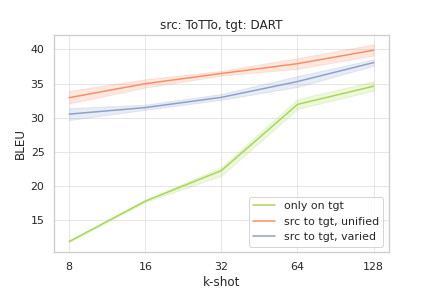}
         \caption{\texttt{ToTTo (table) to DART (KG)}: \textbf{BLEU}}
         \label{fig:totto-to-dart-bleu}
     \end{subfigure}
     \hfill
     \begin{subfigure}{0.3\textwidth}
         \centering
         \includegraphics[width=\textwidth]{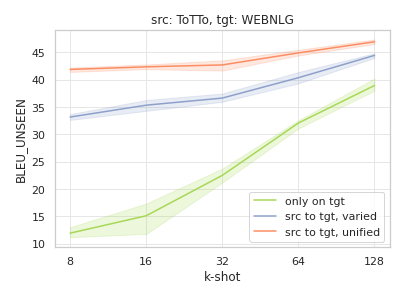}
         \caption{\texttt{ToTTo (table) to WebNLG (KG)}: \textbf{BLEU (Unseen)}}
         \label{fig:totto-to-webnlg-bleu-unseen-main}
     \end{subfigure}
     \hfill
     \begin{subfigure}{0.3\textwidth}
         \centering
         \includegraphics[width=\textwidth]{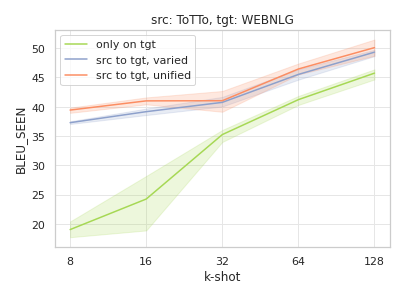}
         \caption{\texttt{ToTTo (table) to WebNLG (KG)}: \textbf{BLEU (Seen)}}
         \label{fig:totto-to-webnlg-bleu-seen-main}
     \end{subfigure}
     
      \begin{subfigure}{0.3\textwidth}
         \centering
         \includegraphics[width=\textwidth]{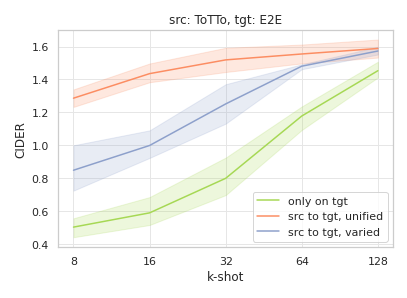}
         \caption{\texttt{ToTTo (table) to E2E (MR)}: \textbf{CIDEr}}
         \label{fig:totto-to-e2e-cider-main}
     \end{subfigure}
     \hfill
     \begin{subfigure}{0.3\textwidth}
         \centering
         \includegraphics[width=\textwidth]{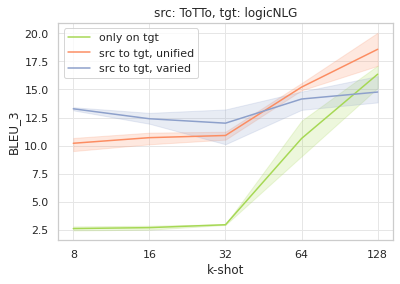}
         \caption{\texttt{ToTTo to LogicNLG (table)}: \textbf{BLEU}}
         \label{fig:totto-to-logicnlg-bleu}
     \end{subfigure}
     \hfill
     \begin{subfigure}{0.3\textwidth}
         \centering
         \includegraphics[width=\textwidth]{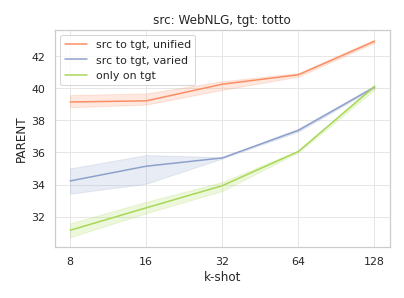}
         \caption{\texttt{WebNLG (KG) to ToTTo (table)}: \textbf{PARENT}}
         \label{fig:wenblg-to-totto-parent-main}
     \end{subfigure}
     
    \caption{Results of few-shot experiments transferring models between two structured forms. Each figure shows three curves, the green curve \textit{"only on tgt"} is the performance of the T5-base model fine-tuned directly on the target task, the red curve \textit{"src to tgt, unified"} is the performance of the model fine-tuned on both tasks using our proposed unified representation, and the blue curve \textit{"src to tgt, varied"} is the performance of the model fine-tuned on both tasks using linearization from other papers, resulting in varied linearization for source and target tasks. The LogicNLG task differs from ToTTo by requiring the model to generate insights from analyzing a table rather than generating descriptions from highlighted cells.}
    \label{fig:totto-to-dart}
\end{figure*}

\begin{table*}[hbt]
\adjustbox{max width=\textwidth}{%
\centering
\footnotesize
\scalebox{0.9}{
\begin{tabular}{l|c|cc|c|ccc|c|cc}
\toprule

    \multirow{2}{*}{\textbf{Model}}  & \multirow{2}{*}{\textbf{Linear}} & \multicolumn{2}{c|}{ \textbf{ToTTo}} & \multicolumn{1}{c|}{ \textbf{DART}}  & \multicolumn{3}{c|}{ \textbf{WebNLG}} & \multicolumn{1}{c|}{ \textbf{LogicNLG}} & \multicolumn{2}{c}{ \textbf{E2E}} \\
    &&  \textbf{BLEU} & \textbf{PARENT} & \textbf{BLEU}  & \textbf{S} & \textbf{U} & \textbf{A} & \textbf{BLEU-3} & \textbf{BLEU} & \textbf{CIDEr} \\
\midrule
    \multicolumn{11}{c}{\textit{Single-Task Training}} \\
\midrule
    \textit{LATTICE}\citep{wang-etal-2022-robust}  & Tab & 48.6 & -  & - & - & - & - & 20.1 & -  & - \\
    \textit{UnifiedSKG (base)}   & O            & 48.3 & -  & 46.2  & - & - & - &-  & - & - \\
    \textit{UnifiedSKG (3B)}    & O            & 49.0 & -  & 46.7  & - & - & - & - & - & - \\
    \textit{DCVED}\citep{chen-etal-2021-de}             & O     &-&-&-&-&-&- &  15.3 &  - & -   \\
    \textit{HTLM}\citep{aghajanyan2022htlm}               & O     &-&- & 47.2 & 65.4 & 48.4 & 55.6 & - & - & - \\
\midrule
    \multirow{2}{*}{T5-base}    & Uni        & 49.3 & 58.9 & 48.6  & 65.4 & 50.1 & 58.5 & 24.7 &  41.8 & 1.90 \\ 
     &         O            & 49.2 & 58.9 & 49.0  & \textbf{65.9} & 49.5 & 58.2 & 25.2  & 42.1 & 1.91 \\ 
\midrule
    \multirow{2}{*}{T5-3B}    & Uni & 49.4 & 58.9 & \textbf{49.6}  & 65.1 & 52.7 & 59.5 & 25.1  & \textbf{42.8} & 1.92 \\
                              & O     & \textbf{49.6} & \textbf{59.0} & 49.3  & 65.3 & \textbf{53.5} & \textbf{60.0} & \textbf{25.3}  & 42.5 & \textbf{1.94} \\
\midrule
    \multicolumn{11}{c}{\textit{Multi-Task Training}} \\
\midrule
    \textit{UnifiedSKG (base)}       & O         & 45.3 & -  & -  & - & - & -  & - & - & - \\
    \textit{C-P} (large) \citep{control-prefix}  & O     &  -   & -    & 52.0 &  67.0 & 55.6 & 61.8 & - & 44.2 & -\\
\midrule
    \multirow{2}{*}{T5-base}   & Uni        & 49.7 & 59.2 & 49.8  & 64.9 & 50.3 & 58.3 & 25.2  & 42.9 & 1.94 \\ 
     &                        O         & 48.5 & 58.7 & 48.1  & 64.1 & 50.2 & 57.9 & 24.7 & 41.7 & 1.89 \\
\midrule
    \multirow{2}{*}{T5-3B}  & Uni & \textbf{50.8} & \textbf{60.4} &  \textbf{50.2}  & \textbf{65.4} & \textbf{53.4} & \textbf{60.0} & \textbf{25.4}  & \textbf{43.2} & \textbf{1.99} \\
     &                     O & 50.2 & 59.5 & 49.8  & 65.3 & 51.9 & 59.4 & 25.3  & 41.8 & 1.89 \\
\bottomrule
\end{tabular}}
}
\caption{Single-task and multi-task training results using full training sets. In the "Linear" column, "Uni" represents using unified representation, "O" means using other linearizations from previous papers, and "Tab" mean we use table-specific encoder.}
\label{tab:all}
\end{table*}


\subsubsection{Zero-Shot Performance}
\label{sssec:zero-shot}
The zero-shot results are summarized in Table \ref{tab:zero-shot}. We compare our results to recent works GPT2-XL \citep{Keymanesh2022WhatMD}, KGPT \citep{chen-etal-2020-kgpt}, JointGT \citep{ke-etal-2021-jointgt} and HTLM \citep{aghajanyan2022htlm}. Both KGPT and JointGT models are pretrained on large amounts of aligned knowledge graph and text data. HTLM is a hyper-text language model pre-trained on a large-scale web crawl. It allows for structured prompting in the HTML format.

From the results, we make several observations. \textbf{(1)} The \textit{Only on tgt} performance is very low as expected, as the T5-base model has not been trained on any data. However, surprisingly the NLI score on LogicNLG is the highest under this setting. We observe that this NLI score is very unstable and might not be a good metric for judging the entailment of generated text. \textbf{(2)} The performance of \textit{Src to tgt, unified} consistently and significantly surpasses that of \textit{Src to tgt, varied}, even though both models are trained using the same source-task data, but with different representations. This demonstrates that representing source and target tasks in the same format is crucial for successful zero-shot transfer, as a common representation facilitates the transfer of knowledge learned on the source data to other structured forms and tasks. \textbf{(3)} The zero-shot performance of the "unified" model is even better than few-shot results of the baseline models. On DART, the "unified" model's BLEU score is 43\% higher than that of HTLM. The improvement on WebNLG is particularly noteworthy for unseen categories. Utilizing a unified representation results in a zero-shot BLEU score of 39.82, surpassing the few-shot results of 37.18 by \citet{ke-etal-2021-jointgt} and 18.5 by \citet{aghajanyan2022htlm}.


\subsubsection{Few-Shot Results}
\label{sssec:few-shot}
Figure \ref{fig:totto-to-dart} shows the few-shot results for sample sizes 8, 16, 32, 64, and 128. We repeat the experiments 5 times for each sample size and report the mean and 95\% confidence intervals.

\paragraph{Table $\longrightarrow$ KG Triples}
From Figure \ref{fig:totto-to-dart-bleu}, \ref{fig:totto-to-webnlg-bleu-unseen-main} and \ref{fig:totto-to-webnlg-bleu-seen-main}, we have identified three key observations: (1) Both the models \textit{Src to tgt, unified} and \textit{Src to tgt, varied}, which were initially trained on ToTTo, perform significantly better than the model \textit{Only on tgt}, which was only trained on target tasks. This indicates that these two structured forms share common knowledge and that training the model on tabular input can greatly enhance its understanding of KG triples. (2) Furthermore, \textit{Src to tgt, unified} (represented by the red curve) outperforms \textit{Src to tgt, varied} (represented by the blue curve) by a substantial margin. This observation aligns with our previous findings in the zero-shot setting (as seen in Table \ref{tab:zero-shot}) and highlights the importance of our unified representation approach in transferring knowledge learned from tables to KG triples. (3) Additionally, on the task of WebNLG, the improvement on unseen categories is particularly notable, further reinforcing our zero-shot findings.

\paragraph{Table $\longrightarrow$ Meaning Representations}
Based on Figure \ref{fig:totto-to-e2e-cider-main}, similar observations can be made for the E2E dataset. 
The improvement in terms of CIDEr is particularly significant when using fewer than 64 samples, indicating that the unified model generates more informative text compared to the varied and vanilla models.

\paragraph{Table Description $\longrightarrow$ Table Insights}
The LogicNLG task is distinct from the ToTTo task in that it requires the model to generate insights by analyzing the contents of a table, rather than generating surface-form descriptions based on highlighted cells. As shown in Figure \ref{fig:totto-to-logicnlg-bleu}, when using only 8 samples, the \textit{Src to tgt, varied} model performs better than the \textit{Src to tgt, unified} model. This may be due to the fact that both tasks involve generating text from tables, and that the unified model is more proficient at transferring knowledge learned on the source task to the target task, which may lead to the generation of table descriptions rather than insights when provided with a limited number of samples. However, as the number of samples increases, the performance of the unified model improves, and it surpasses the varied model when k=128. A concrete example is provided in the case study section \ref{sec:case_study} to further illustrate our observation.

\paragraph{KG Triples $\longrightarrow$ Table}
The benefits of utilizing unified representation are particularly substantial when transferring models that have been trained on knowledge graphs to table inputs. In Figure \ref{fig:wenblg-to-totto-parent-main}, the PARENT gap between unified and varied models is consistently greater than 2 points. In fact, the performance of "varied" and "only on tgt" models converge when utilizing 128 samples, and is only slightly superior to that of the "unified" model when provided with only 8 samples. This suggests that the use of unified representation is highly efficient in terms of sample utilization.

\subsection{Full-Set Finetuning Results}
In this section, we train the models on full training sets, in either single-task or multi-task settings. Additional experimental results are presented in the appendix.

\paragraph{Single-Task Training}
\label{sssec:single-task}
From the "single-task training" results in Table \ref{tab:all}, a key finding is that the proposed unified representation method results in performance comparable to other linearization techniques studied in previous research. This is particularly evident on the DART, WebNLG, and E2E tasks, where the data was first converted into virtual tables, and the results from both methods are similar, indicating that this conversion does not result in a significant loss of information.




\paragraph{Multi-Task Training}
\label{sssec:multi-task}
The performance of multi-task models is summarized in Table \ref{tab:all} under the "multi-task training" section, revealing several key findings:
\textbf{(1)} \textit{Overall, multi-task training using different linearizations for each dataset results in a \textbf{worse} performance compared to single-task training.} BLEU scores for T5-base models decrease from 49.2 to 48.5 on ToTTo, from 49.0 to 48.1 on DART, and from 65.9 to 64.1 on seen categories of WebNLG. This confirms the findings of UnifiedSKG \citep{UnifiedSKG}, which found that single-task model performance was higher than multi-task performance on ToTTo dataset. However, it is unclear if this drop in performance was due to task differences, as their study included other tasks. Our results provide further insight into data-to-text tasks alone and show that multi-task performance can still be inferior if input formats are not unified.
\textbf{(2)} In contrast, \textit{multi-task trained "unified" models consistently outperform single-task models}, with the only exception of the base model on the WebNLG dataset. This demonstrates that utilizing a unified representation approach helps models learn common knowledge across various tasks without negatively impacting performance.
\textbf{(3)} \textit{The "unified" models consistently demonstrate superior performance compared to "varied" models in multi-task training}, with a larger margin of improvement observed in base-sized models.




\subsection{Qualitative Study}
\label{sec:case_study}

\begin{table*}[htb]
\adjustbox{max width=\textwidth}{%
\centering
\footnotesize
\scalebox{0.9}{
\begin{tabular}{p{0.1\linewidth} | p{0.44\linewidth} | p{0.44\linewidth}}
\toprule
\textbf{k-shot =} & \textbf{Src to tgt, unified} & \textbf{Src to tgt, varied} \\
\midrule
\multicolumn{3}{c}{\textit{ToTTo (table) $\longrightarrow$ WebNLG (KG) example}} \\
\midrule
8 & Hip hop music is influenced by Disco by Allen Forrest (born in Fort Campbell) and Funk with drum and bass. & Allen Forrest was born in Fort Campbell. \\
\midrule
128 & Allen Forrest, born in Fort Campbell, is known for his roots in hip hop music. \hlc[green]{Disco and Funk} are stylistic origins, while drum and bass are derivatives. & Allen Forrest was born in Fort Campbell and is known for hip hop music. Hip hop music is a derivative of drum and bass. \\
\midrule
\textbf{Groundtruth} &  \multicolumn{2}{p{0.88\linewidth}}{Allen Forrest was born in Fort Campbell and is a hip hop musician. Hip hop originates from \hlc[green]{funk and disco} and was derived into drum and bass music.} \\
\midrule
\textbf{KG triples} & \multicolumn{2}{p{0.88\linewidth}}{(\texttt{Hip hop music, stylistic origin, \hlc[green]{Disco}) (Allen Forrest, birth place, Fort Campbell) (Allen Forrest, genre, Hip hop music) (Hip hop music, stylistic origin, \hlc[green]{Funk}) (Hip hop music, derivative, Drum and bass)}} \\
\midrule
\multicolumn{3}{c}{\textit{ToTTo (table) $\longrightarrow$ E2E (MR) example}} \\
\midrule
8 & Zizzi is a pub near The Sorrento. & Zizzi is a gastropub in Sorrento, near The Sorrento. \\
\midrule
128 & Zizzi is a pub near The Sorrento. & Zizzi is a pub near The Sorrento. \\
\midrule
\textbf{Groundtruth} &  \multicolumn{2}{p{0.88\linewidth}}{There is a pub called Zizzi located near The Sorrento.} \\
\midrule
\textbf{MRs} & \multicolumn{2}{p{0.88\linewidth}}{\texttt{name[Zizzi], eatType[pub], near[The Sorrento]}} \\
\midrule
\multicolumn{3}{c}{\textit{ToTTo (table) $\longrightarrow$ LogicNLG (table) example}} \\
\midrule
8 & In the world golf championships, the United States has 12 individual winners, Australia has 3 individual winners, England has 3 individual winners, South Africa has 1 individual winner, Canada has 1 individual winner, Fiji has 1 individual winner, Italy has 1 individual winner, Japan has 0 individual winner, and Wales has no individual winner. & The United States has the highest number of individual winners of any country in the world. \\
\midrule
128 & The United States is the only nation to have won 12 World Golf Championship. & The United States has the highest number of individual winners. \\
\midrule
\textbf{Input table} & \multicolumn{2}{p{0.99\linewidth}}{\includegraphics[width=\linewidth]{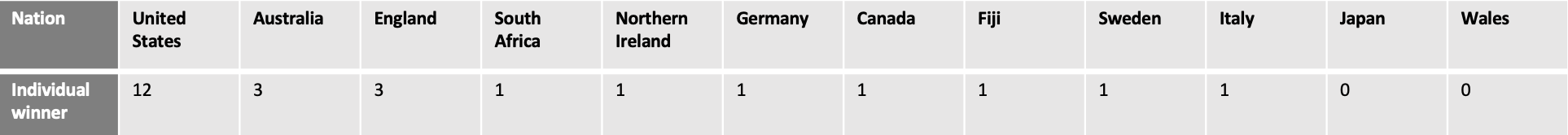}} \\
\bottomrule
\end{tabular}}
}
\caption{Case study: few-shot prediction examples on WebNLG and E2E.}
\label{tab:webnlg-examples}
\end{table*}

We conduct a qualitative case study to compare the texts generated by the \textit{Src to tgt, unified} and \textit{Src to tgt, varied} models. The results are illustrated in Table \ref{tab:webnlg-examples}, which displays the model's generations for different sample sizes.

For the WebNLG example, the input contains 5 KG triples. When $k=8$, the "varied" model only covers one KG triple fact, while the "unified" model includes many more nodes and relations from the input. As the sample size increases to 128, the "unified" model's generation covers all facts accurately, while the "varied" model's generation still misses the "funk and disco" origin of pop music.

In the E2E example, the "unified" model output is consistent and accurate with both 8 and 128 samples. In contrast, the "varied" model produces "Sorrento" twice. This serves as additional evidence that using a unified representation enhances the transfer of the generation style learned on table input to meaning representations.

The results of the LogicNLG input generation offer validation for our hypothesis that the "unified" model performs less effectively than the "varied" model when the sample size is small, due to its persistent focus on generating descriptions of the table input, as it has been trained to do on the ToTTo data. Indeed, the descriptions generated by the "unified" model when sample size is 8, are accurate reflections of the table's content. When the sample size is increased to 128, both models generate sentences that are more akin to insights. It is noteworthy that the "unified" model generates "world golf championship" even though it is not present in the table, which pertains to the golf championship. We posit that this information is carried over from the ToTTo data, and the "unified" model is able to retain this information while the "varied" model does not.

\section{Conclusion and Future Work}
\label{sec:conclusion}

We have introduced a unified representation approach for data-to-text tasks, which effectively converts table contents, knowledge graph triples, and meaning representations into a single representation. Our experiments demonstrate that this unified representation significantly improves generalization across different structured forms, especially in zero-shot or few-shot settings. Our method is particularly beneficial in situations where data is scarce. Additionally, by using the unified representation, our multi-task-trained models consistently outperform single-task models, which is in contrast to previous findings that mixing different data types can negatively impact overall performance.

One future direction is to apply our method to other tasks that involve heterogeneous inputs, such as question answering over knowledge bases, where knowledge can be stored in both tables and knowledge graphs. It would also be interesting to investigate whether a model pre-trained on large knowledge graphs can more effectively transfer learned commonsense knowledge to table QA tasks, when using our unified representation approach.

\section*{Limitations}
It is important to note that the unified representation proposed in our study is just one option among many. Other linearization methods may potentially yield better results. For example, research by \citet{Yin2022NaturalLT} and \citet{aghajanyan2022htlm} has explored using code generation with Jupyter notebooks and a hyper-text language model with structured prompting, respectively. Further research in these areas, such as converting all structured forms to markdown language or hyper-texts, may yield alternative unified representations.

\section*{Ethics Statement}
We acknowledge the importance of the ACL Ethics Policy and agree with it. This study addresses the problem of data-to-text generation and explores whether a unified representation can enhance cross-task performance on various structured forms. Since our input comes from knowledge bases, a potential concern is that biases or fairness issues may be present in the KB, which could also be reflected in the generated text. Therefore, it is crucial to use the model with caution in practice. We believe this work can contribute to the field of data-to-text generation, particularly in situations where data is scarce.


\bibliography{custom}
\bibliographystyle{acl_natbib}

\clearpage
\newpage
\appendix



\section{Data Statistics}
We summarize the input type and number of examples in each dataset.
\begin{table}[htb]
\footnotesize
\centering
\scalebox{0.9}{
\begin{tabular}{l|c|ccc}
\toprule

    \multirow{2}{*}{\textbf{Dataset}} & \multirow{2}{*}{\textbf{Input}}
    & \multicolumn{3}{c}{ \textbf{\# Examples}}\\
    &&  \textbf{Train} & \textbf{Validation} &\textbf{Test} \\
\midrule
    ToTTo	  & Table    & 120,761   & 7,700	  &7,700\\	
    DART      & KG	& 30,526 & 2,768  & 6,959 \\ 
    LogicNLG  & Table	& 28,450  & 4,260  & 4,305 \\
    WebNLG & KG	&  18,102 & 872 & 1,862 \\ 
    E2E clean    & MR &  33,525 & 1,484 & 1,847 \\
\bottomrule
\end{tabular}}
\caption{Data statistics.}
\label{tab:data}
\end{table}

\section{Column-wise Linearization of (Sub)Table}
\label{col-wise-example}
A column-wise linearization of Table \ref{tab:example} is:

\adjustbox{max width=\textwidth}{%
\begin{lstlisting}[language=Html, moredelim={[is][\color{magenta}]{START}{END}}]
<title> Alma Jodorowsky </title>
<STARTsub\_titleEND> Filmography <START/sub\_titleEND>
<table>
  <STARTcolumnEND>
    <STARTcol_headerEND> Year <START/col_headerEND>
      <STARTcellEND> 2014 <START/cellEND>
      <STARTcellEND> 2016 <START/cellEND>
      <STARTcellEND> 2017 <START/cellEND>
  <START/columnEND>
  ... other columns ...
</table>
\end{lstlisting}
}

\section{Other Linearizations Used in Previous Papers}
\paragraph{Table highlights}: Our unified representation is motivated by ToTTo linearization, and hence they are very similar. The only difference is ToTTo uses \texttt{<page\_title>} instead of \texttt{<title>} and \texttt{<section\_title>} instead of \texttt{<sub\_title>}.

\paragraph{KG triples}: Given a set of triples \texttt{\{(William Wasmund, FIELD\_GOALS,
0), (William Wasmund, EXTRA\_POINTS, 0)\}}, an alternative linearization used in UnifiedSKG \citep{UnifiedSKG} is \texttt{William Wasmund : field goals : 0 | William Wasmund : extra points : 0}

\paragraph{Entire table}: The alternative linearization used in LogicNLG \citep{chen-etal-2020-logical} for Table \ref{tab:example} is: \texttt{Given the table title of Alma Jodorowsky, Filmograph. In row 1 , the Year is 2014 , the Title is La ..., the Role is Solana, the Notes is TV ... In row 2 , ...}

\paragraph{Mearning representation}: The alternative linearization we use for the example in Figure \ref{fig:method} is simply concatenating all the MRs: \texttt{name[Cocum], eatType[coffee shop], food[Italian], priceRange[cheap], familyFriendly[yes]}.\\

\section{Implementation Details}
In the zero- and few-shot experiments, we employ the T5-base model as the base model and train it for 30 epochs for both the source and target tasks. For the source task, we use a learning rate of 5e-5 and a batch size of 32, and for the target task, we use a learning rate of 2e-5 and a batch size of 8.

\section{More Multi-Task Results}
\label{sec:appendix-e2e}
We present more detailed multi-task results on each of the dataset in this section. The results are summarized in Table \ref{tab:e2e-appendix}, \ref{tab:totto-appendix}, \ref{tab:dart-appendix} and \ref{tab:logicnlg-appendix}.

\begin{table*}[htb]
\centering
\footnotesize
\scalebox{0.9}{
\begin{tabular}{l|c|c|ccccc}
\toprule

    \textbf{Model} & \textbf{Task} & \textbf{Linearization} & \textbf{METEOR} & \textbf{ROUGE-L} & \textbf{CIDEr} & \textbf{NIST} & \textbf{BLEU} \\
\midrule
    \textsc{Control Prefix} (large)  & MT & Alt & 39.2 & - & - & - & 44.2 \\
\midrule
    \multirow{4}{*}{T5-base}  & ST & Unified & 38.3 & 56.6 & 1.90 & 6.20 & 41.8 \\ 
       & ST & Alt     & 38.3 & 56.4 & 1.91 & 6.23 & 42.1 \\ 
       & MT & Unified & \textbf{38.6} & \textbf{57.0} & \textbf{1.94} & \textbf{6.31} & \textbf{42.9} \\ 
       & MT & Varied  & 38.3 & 56.6 & 1.89 & 6.20 & 41.7 \\
\midrule
    \multirow{4}{*}{T5-3B}    & ST & Unified & 38.5 & 56.7 & 1.92 & 6.30 & 42.8 \\
         & ST & Alt     & 38.5 & 56.5 & 1.94 & 6.31 & 42.5 \\
         & MT & Unified & \textbf{38.7} & \textbf{57.4} & \textbf{1.99} & \textbf{6.34} & \textbf{43.2} \\
         & MT & Varied  & 38.3 & 56.8 & 1.89 & 6.21 & 41.8 \\
\bottomrule
\end{tabular}}
\caption{Test set performance on E2E clean.}
\label{tab:e2e-appendix}
\end{table*}

\begin{table*}[htb]
\centering
\footnotesize
\scalebox{0.9}{
\begin{tabular}{l|c|c|cc|cc|cc}
\toprule
    \multirow{2}{*}{\textbf{Model}} & \multirow{2}{*}{\textbf{Task}} & \multirow{2}{*}{\textbf{Linearization}} & \multicolumn{2}{c}{ \textbf{Overall}}  & \multicolumn{2}{c}{ \textbf{Overlap}} & \multicolumn{2}{c}{\textbf{Non-overlap}} \\
    &&&  \textbf{BLEU} & \textbf{PARENT} & \textbf{BLEU} & \textbf{PARENT} & \textbf{BLEU} & \textbf{PARENT} \\
\midrule
    \textit{LATTICE (T5-base)}  & ST & Table-specific & 48.6 & -    & 56.6 &  -    & 40.8 &   -   \\
    \textit{UnifiedSKG (base)}  & ST & Alt           & 48.3 & -    & -    & -    & -    &  -    \\
    \textit{UnifiedSKG (base)}  &  MT & Varied           & 45.3 & -    & -    & -    & -    &  -    \\
    \textit{UnifiedSKG (3B)}    & ST & Alt           & 49.0 & -    & -    & -    & -    &   -   \\
    \textit{Text2Text (3B)}           & ST & Alt           & 48.4 & 57.8 & -    & -    & 40.4 & 53.3 \\
\midrule
    \multirow{3}{*}{T5-base}  & ST & Unified & 49.3 & 58.9 & 57.1 & 62.7 & \textbf{41.9} & \textbf{55.3} \\ 
     &  MT & Unified & \textbf{49.7} & \textbf{59.2} & \textbf{57.7} & \textbf{63.2} & \textbf{41.9} & 55.2 \\ 
     &  MT & Varied & 48.5 & 58.7 & 56.2 & 62.5 & 41.1 & 55.0 \\
\midrule
    \multirow{3}{*}{T5-3B}    & ST & Unified & 49.4 & 58.9 & 57.1 & 62.7 & 42.0 & 55.3 \\
     &  MT & Unified & \textbf{50.8} & \textbf{60.4} & \textbf{58.5} & \textbf{64.4} & \textbf{43.4} & \textbf{56.5} \\
     &  MT & Varied & 50.2 & 59.5 & 57.5 & 63.2 & 43.2 & 55.9 \\
\bottomrule
\end{tabular}}
\caption{Development set performance on ToTTo.}
\label{tab:totto-appendix}
\end{table*}

\begin{table*}[htb]
\centering
\footnotesize
\scalebox{0.9}{
\begin{tabular}{l|c|c|ccc|ccc}
\toprule
    \multirow{2}{*}{\textbf{Model}} & \multirow{2}{*}{\textbf{Task}} & \multirow{2}{*}{\textbf{Linearization}} & \multicolumn{3}{c}{ \textbf{DART}}  & \multicolumn{3}{c}{ \textbf{WebNLG}}  \\
    &&& \textbf{BLEU} ($\uparrow$) & \textbf{METERO} ($\uparrow$) & \textbf{TER} ($\downarrow$) & \textbf{Seen} & \textbf{Unseen} & \textbf{All} \\
\midrule
    UnifiedSKG (base)  & ST & Alt             & 46.2 & -    & -   &  - & - & -    \\
    UnifiedSKG (3B)    & ST & Alt             & 46.7 & -    & -   &  - & - & -    \\
    \textsc{Control Prefix} (large) &  MT & Alt &52.0 & 0.41 & 0.43 &  67.0 & 55.6 & 61.8 \\
\midrule
    \multirow{4}{*}{T5-base}  & ST & Unified &  48.6 & \textbf{0.40} & 0.45 & 65.4 & 50.1 & \textbf{58.5} \\ 
      & ST & Alt     &  49.0 & \textbf{0.40} & 0.45 & \textbf{65.9} & 49.5 & 58.2 \\ 
    &  MT & Unified &  \textbf{49.8} & \textbf{0.40} & \textbf{0.44} & 64.9 & \textbf{50.3} & 58.3 \\ 
    &  MT & Varied  &  48.1 & 0.39 & 0.45 & 64.1 & 50.2 & 57.9 \\
\midrule
    \multirow{4}{*}{T5-3B}    & ST & Unified &  49.6 & 0.40 & 0.45 & 65.1 & 52.7 & 59.5 \\
        & ST & Alt     &  49.3 & 0.40 & 0.45 & 65.3 & \textbf{53.5} & \textbf{60.0} \\
    &  MT & Unified &  \textbf{50.2} & 0.40 & \textbf{0.44} & \textbf{65.4} & 53.4 & \textbf{60.0} \\
    &  MT & Varied  &  49.8 & 0.40 & \textbf{0.44} & 65.3 & 51.9 & 59.4 \\
\bottomrule
\end{tabular}}
\caption{Test set performance on DART and WebNLG.}
\label{tab:dart-appendix}
\end{table*}

\begin{table*}[htb]
\footnotesize
\centering
\scalebox{0.9}{
\begin{tabular}{l|c|c|c|ccc|cc}
\toprule

    \multirow{2}{*}{\textbf{Model}} & \multirow{2}{*}{\textbf{Task}} & \multirow{2}{*}{\textbf{Linearization}} & \multirow{2}{*}{\textbf{Orientation}} & \multicolumn{3}{c}{ \textbf{Surface-Level Fidelity}} & \multicolumn{2}{c}{ \textbf{Logical Fidelity}}  \\
    & & & & \textbf{BLEU-1}  & \textbf{BLEU-2}  & \textbf{BLEU-3} & \textbf{NLI-acc} & \textbf{SP-acc} \\
\midrule
    GPT-TabGen         & ST & Alt   & row        & 48.8 & 27.1    & 12.6 & 68.7 & 42.1   \\
    DCVED              & ST & Alt  & row         & 49.5 & 28.6    & 15.3 & 76.9 & 43.9   \\
\midrule
    \multirow{5}{*}{T5-base}  & ST & Unified &  column & 52.8 & 34.9 & 24.3 & 79.6 & 45.2 \\ 
      & ST & Unified &  row    & 53.3 & 35.4 & 24.7 & 84.7 & 45.8  \\ 
      & ST & Alt     &  row    & \textbf{54.6} & \textbf{36.1} & \textbf{25.2} & \textbf{85.5} & 45.9 \\ 
      &  MT & Unified &  column & 53.8 & 35.8 & 25.1 & 78.7 & \textbf{47.2}  \\ 
      &  MT & Unified &  row    & 54.4 & \textbf{36.1} & \textbf{25.2} & 80.4 & 46.3 \\ 
      &  MT & Varied  &  row    & 53.9 & 35.5 & 24.7 & 84.2 & 46.3 \\
\midrule
    \multirow{5}{*}{T5-3B}    & ST & Unified &  column & 54.9 & 36.4 & 25.4 & \textbf{88.4} & \textbf{49.8} \\
        & ST & Unified &  row    & 54.1 & 35.9 & 25.1 & 87.1 & 47.9 \\
        & ST & Alt     &  row    & 54.4 & 36.1 & 25.3 & 81.1 & 47.3 \\
        &  MT & Unified &  column & 54.8 & 36.3 & \textbf{25.4} & 87.0 & 49.4 \\
        &  MT & Unified &  row    & \textbf{55.1} & \textbf{36.4} & \textbf{25.4} & 82.9 & 49.1 \\
        &  MT & Varied  &  row    & 54.4 & 36.0 & 25.3 & 80.7 & 47.4 \\
\bottomrule
\end{tabular}}
\caption{Test set performance on LogicNLG.}
\label{tab:logicnlg-appendix}
\end{table*}

\section{More Few-shot Results}
\begin{figure*}[htb]
     \centering
     \begin{subfigure}{0.3\textwidth}
         \centering
         \includegraphics[width=\textwidth]{4_experiments/totto-to-webnlg-bleu_unseen.png}
         \label{fig:totto-to-webnlg-bleu-unseen}
     \end{subfigure}
     \hfill
     \begin{subfigure}{0.3\textwidth}
         \centering
         \includegraphics[width=\textwidth]{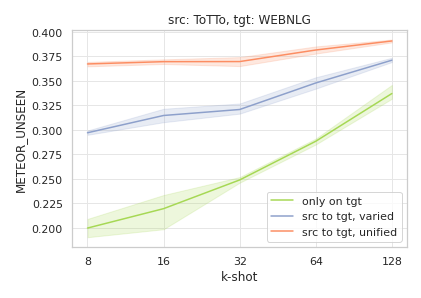}
         \label{fig:totto-to-webnlg-meteor-unseen}
     \end{subfigure}
     \hfill
     \begin{subfigure}{0.3\textwidth}
         \centering
         \includegraphics[width=\textwidth]{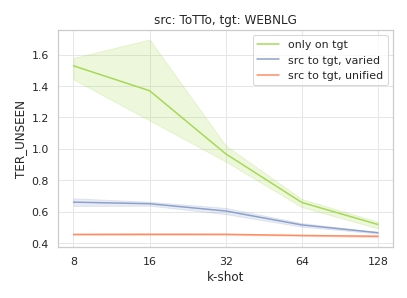}
         \label{fig:totto-to-webnlg-ter-unseen}
     \end{subfigure}
     \begin{subfigure}{0.3\textwidth}
         \centering
         \includegraphics[width=\textwidth]{4_experiments/totto-to-webnlg-bleu_seen.png}
         \label{fig:totto-to-webnlg-bleu-seen}
     \end{subfigure}
     \hfill
     \begin{subfigure}{0.3\textwidth}
         \centering
         \includegraphics[width=\textwidth]{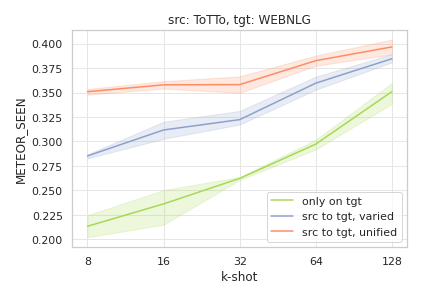}
         \label{fig:totto-to-webnlg-meteor-seen}
     \end{subfigure}
     \hfill
     \begin{subfigure}{0.3\textwidth}
         \centering
         \includegraphics[width=\textwidth]{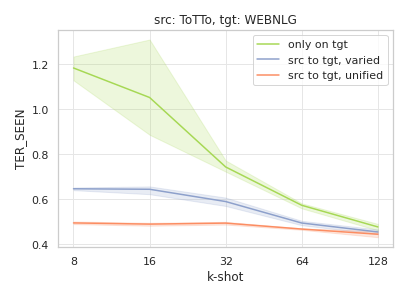}
         \label{fig:totto-to-webnlg-ter-seen}
     \end{subfigure}
     
    \caption{Few-shot experiments of format transferring from ToTTo (table) to WebNLG (KG triples).}
        \label{fig:totto-to-webnlg-appendix}
\end{figure*}

\begin{figure*}[htb]
     \centering
     \begin{subfigure}{0.3\textwidth}
         \centering
         \includegraphics[width=\textwidth]{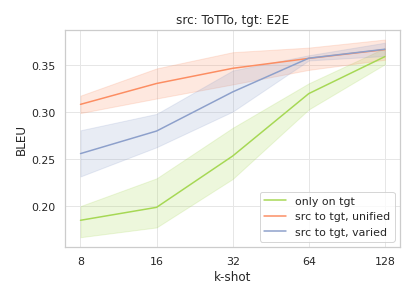}
         \caption{BLEU}
         \label{fig:totto-to-e2e-bleu}
     \end{subfigure}
     \begin{subfigure}{0.3\textwidth}
         \centering
         \includegraphics[width=\textwidth]{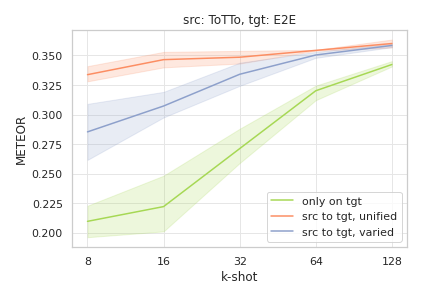}
         \caption{METEOR}
         \label{fig:totto-to-e2e-meteor}
     \end{subfigure}
    \begin{subfigure}{0.3\textwidth}
         \centering
         \includegraphics[width=\textwidth]{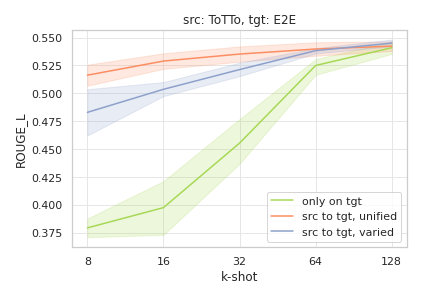}
         \caption{ROUGE-L}
         \label{fig:totto-to-e2e-rouge}
     \end{subfigure}
    \begin{subfigure}{0.3\textwidth}
         \centering
         \includegraphics[width=\textwidth]{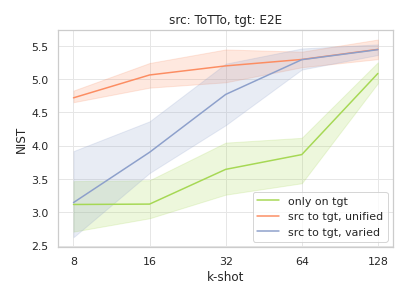}
         \caption{NIST}
         \label{fig:totto-to-e2e-nist}
     \end{subfigure}
     \begin{subfigure}{0.3\textwidth}
         \centering
         \includegraphics[width=\textwidth]{4_experiments/totto-to-e2e-CIDEr.png}
         \caption{CIDEr}
         \label{fig:totto-to-e2e-cider}
     \end{subfigure}
     
    \caption{Few-shot experiment results of task transferring from ToTTo (table) to E2E (meaning representation).}
        \label{fig:totto-to-e2e-appendix}
\end{figure*}

\begin{figure*}[htb]
     \centering
     \begin{subfigure}{0.3\linewidth}
         \centering
         \includegraphics[width=\textwidth]{4_experiments/totto-to-logicnlg-bleu-3.png}
     \end{subfigure}
     \begin{subfigure}{0.3\linewidth}
         \centering
         \includegraphics[width=\textwidth]{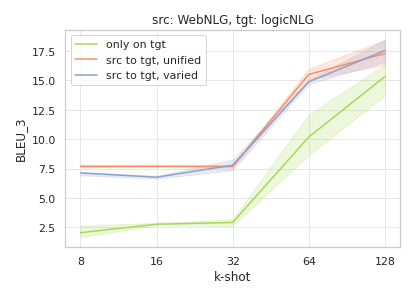}
     \end{subfigure}
     
    \caption{Few-shot experiment results of task transferring from ToTTo (table highlights description) or WebNLG (KG triples) to LogicNLG (logical inference on table).}
        \label{fig:totto-to-logicnlg-appendix}
\end{figure*}

We present other few-shot results using more metrics in Figure \ref{fig:totto-to-webnlg-appendix}, \ref{fig:totto-to-e2e-appendix} and \ref{fig:totto-to-logicnlg-appendix}.

\section{Human Evaluation}
We conducted a human evaluation on the few-shot ToTTo to WebNLG transferring experiment. Specifically, we randomly selected 50 WebNLG test data from the unseen schema and compared the performance of the 8-shot \textit{src to tgt, unified} and \textit{src to tgt, varied} models.

For each of the 50 samples, we generated texts using both models and asked three annotators to choose the better option based on factuality, coverage of the triples, and fluency. We received only two annotations for two of the samples as one of the annotators did not respond. For the remaining 48 samples, all three annotators reached a consensus on 21 of them (43.75\%). Out of these 21 samples, the "unified" model received unanimous preference from the annotators in 15 cases (71.43\%). If we consider the majority vote among the three annotators, then 75\% of the results favored the "unified" model. The Fleiss Kappa value, which measures agreement among the three annotators, is around 0.23 (fair agreement).

\section{More Qualitative Study}
We present additional few-shot predictions for models transferred from ToTTo to WebNLG and LogicNLG in Tables \ref{tab:webnlg-examples-appendix} and \ref{tab:logicnlg-examples-appendix}, respectively. We also provide error analysis under each example.

\begin{table*}[htb]
\centering
\footnotesize
\scalebox{0.9}{
\begin{tabular}{p{0.1\linewidth} | p{0.44\linewidth} | p{0.44\linewidth}}
\toprule
\textbf{k-shot =} & \textbf{Src to tgt, unified} & \textbf{Src to tgt, varied} \\
\midrule
\multicolumn{3}{c}{\textit{ToTTo (table) $\longrightarrow$ WebNLG (KG) example}} \\
\midrule
8 & Uruguay is led by Ral Fernando Sendic Rodrguez, who died in Montevideo, and by Daniel Martnez, a Spanish politician. & Ral Fernando Sendic Rodrguez died in Montevideo. \\
\midrule
128 & The leader of Uruguay is Ral Fernando Sendic Rodrguez who died in Montevideo, where Alfredo Zitarrosa died. The leader is Daniel Martnez who speaks Spanish. & Ral Fernando Sendic Rodrguez was the leader of Uruguay and Alfredo Zitarrosa died in Montevideo. Daniel Martnez was a politician who led the country in Spanish. \\
\midrule
\textbf{Groundtruth} &  \multicolumn{2}{p{0.88\linewidth}}{Alfredo Zitarrosa died in Montevideo, Uruguay. Daniel Martinez is a political leader in Montevideo, and Raul Fernando Sendic Rodriguez is a leader in Uruguay, where Spanish is spoken.} \\
\midrule
\textbf{KG triples} & \multicolumn{2}{p{0.88\linewidth}}{(\texttt{Uruguay : leader name : Ral Fernando Sendic Rodrguez | Alfredo Zitarrosa : death place : Montevideo | Montevideo : country : Uruguay | Montevideo : leader name : Daniel Martnez (politician) | Uruguay : language : Spanish language}} \\
\midrule
 \textbf{Error analysis} & \multicolumn{2}{p{0.88\linewidth}}{When sample size is 8, the "unified" model generation contains almost all nodes except Alfredo Zitarrosa, but the "varied" model output only contains one triple.} \\
\midrule
\midrule
8 & Twilight (band) is a black metal band with Aaron Turner, and Old Man Gloom is a death metal band with electric guitar. & Twilight is a black metal music fusion genre. \\
\midrule
128 & Twilight (band) is associated with black metal, and Old Man Gloom is associated with death metal, where Aaron Turner played electric guitar. & Twilight is a genre of black metal music and Aaron Turner plays the electric guitar in Old Man Gloom. Death metal is a genre of black metal music. \\
\midrule
\textbf{Groundtruth} &  \multicolumn{2}{p{0.88\linewidth}}{Aaron Turner is an electric guitar player who has played with the black metal band Twilight and with Old Man Gloom. Death metal is a musical fusion of black metal.} \\
\midrule
\textbf{KG triples} & \multicolumn{2}{p{0.88\linewidth}}{(\texttt{Twilight (band) : genre : Black metal | Aaron Turner : associated band/associated musical artist : Twilight (band) | Aaron Turner : associated band/associated musical artist : Old Man Gloom | Aaron Turner : instrument : Electric guitar | Black metal : music fusion genre : Death metal}} \\
\midrule
 \textbf{Error analysis} & \multicolumn{2}{p{0.88\linewidth}}{In "unified" model output, the Twilight is always recognized correctly as a band, but in varied model output, it is wrongly predicted as a genre.} \\
\midrule
\midrule
8 & Al Anderson (NRBQ band) played guitar in rock music, country music and blues music, while Bhangra (music) is a genre of music fusion. & Al Anderson is a guitarist with the NRBQ band. \\
\midrule
128 & Bhangra is a genre of music fusion, where Al Anderson plays guitar, Country music and Blues. & Al Anderson is an Indian country musician who plays guitar and plays Bhangra (music). Country music is a music fusion genre which is influenced by blues and rock music. \\
\midrule
\textbf{Groundtruth} &  \multicolumn{2}{p{0.88\linewidth}}{Al Anderson plays the guitar for rock band NRBQ. Rock comes from country music which originates from blues. Bhangra music is a musical fusion of rock music.} \\
\midrule
\textbf{KG triples} & \multicolumn{2}{p{0.88\linewidth}}{(\texttt{Rock music : music fusion genre : Bhangra (music) | Al Anderson (NRBQ band) : instrument : Guitar | Al Anderson (NRBQ band) : genre : Rock music | Rock music : stylistic origin : Country music | Country music : stylistic origin : Blues}} \\
\midrule
 \textbf{Error analysis} & \multicolumn{2}{p{0.88\linewidth}}{This is an negative example for "unified" model. When sample size is 8, the unified model generation covers more nodes compared to "varied" model generation. However, when sample size is 128, the "unified" model wrongly generates Anderson to play guitar, country music and Blue. This is a wrong fact. But "varied" model generation is able to be more aligned with the input triples.} \\
\bottomrule
\end{tabular}}
\caption{Case study: few-shot prediction examples on WebNLG.}
\label{tab:webnlg-examples-appendix}
\end{table*}

\begin{table*}[htb]
\adjustbox{max width=\textwidth}{%
\centering
\footnotesize
\scalebox{0.9}{
\begin{tabular}{p{0.12\linewidth} | p{0.44\linewidth} | p{0.44\linewidth}}
\toprule
\textbf{k-shot =} & \textbf{Src to tgt, unified} & \textbf{Src to tgt, varied} \\
\midrule
\multicolumn{3}{c}{\textit{ToTTo (highlighted cells) $\longrightarrow$ LogicNLG (table) example}} \\
\midrule
8 & Eric Turner, Ed King, James Jones, Pio Sagapolutele, Michael Jackson, Raymond Irvin, Shawn Wiggins, Brian Greenfield, Todd Jones, Todd Jones, and Elijah Austin played in the 1991 cleveland browns season. & In the 1991 cleveland browns season, Eric Turner, Ed King, Pio Sagapolutele and Raymond Irvin were the Defensive Tackle. \\
\midrule
128 & Raymond Irvin was the Defensive Back in the 1991 Cleveland Browns season. & Brian Greenfield, Todd Jones and Elijah Austin were the Defensive Tackle in the 1991 Cleveland Browns season. \\
\midrule
\midrule
8 & Eric Turner, Ed King, James Jones, Pio Sagapolutele, Michael Jackson, Raymond Irvin, Shawn Wiggins, Brian Greenfield, Todd Jones, Todd Jones, and Elijah Austin played in the 1991 cleveland browns season. & In the 1991 cleveland browns season, Eric Turner, Ed King, Pio Sagapolutele and Raymond Irvin were the Defensive Tackle. \\
\midrule
128 & Raymond Irvin was the Defensive Back in the 1991 Cleveland Browns season. & Brian Greenfield, Todd Jones and Elijah Austin were the Defensive Tackle in the 1991 Cleveland Browns season. \\
\midrule
\midrule
8 & Eric Turner, Ed King, James Jones, Pio Sagapolutele, Michael Jackson, Raymond Irvin, Shawn Wiggins, Brian Greenfield, Todd Jones, Todd Jones, and Elijah Austin played in the 1991 cleveland browns season. & In the 1991 cleveland browns season, Eric Turner, Ed King, Pio Sagapolutele and Raymond Irvin were the Defensive Tackle. \\
\midrule
128 & Raymond Irvin was the Defensive Back in the 1991 Cleveland Browns season. & Brian Greenfield, Todd Jones and Elijah Austin were the Defensive Tackle in the 1991 Cleveland Browns season. \\
\midrule
\textbf{Groundtruths} &  \multicolumn{2}{p{0.88\linewidth}}{"Raymond Irvin is the second Defensive Back to get drafted",
    "Frank Conover is the third Defensive Tackle to get drafted",
    "Elijah Austin is the last Defensive Tackle to get drafted",
    "Frank Conover has an Overall that is 56 higher than Michael Jackson",
    "Shawn Wiggins is the second Wide Receiver to get drafted"} \\
\midrule
\textbf{Input table} & \multicolumn{2}{p{0.99\linewidth}}{\includegraphics[width=\linewidth]{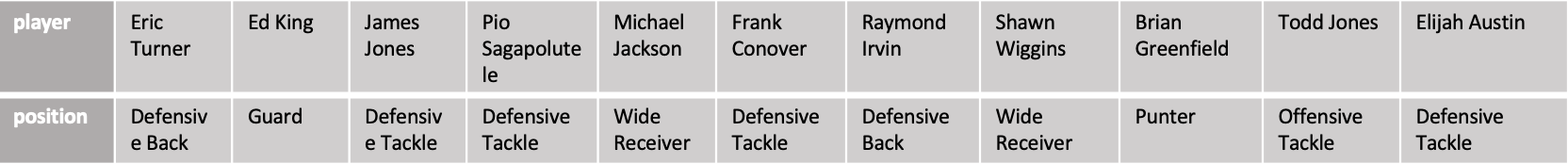}} \\
\midrule
 \textbf{Error analysis} & \multicolumn{2}{p{0.88\linewidth}}{Similar to our analysis in Section \ref{sec:case_study}, the "unified" model generation is more like description when sample size is 8. Again this is because the source task is ToTTo, which is a task to generate surface-level description of table contents. The "unified" model transfers this learned knowledge better, and hence generates sentences that are more like descriptions. When sample size is 128, both models generate similar contents.} \\
\bottomrule
\end{tabular}}
}
\caption{Case study: few-shot prediction examples on LogicNLG.}
\label{tab:logicnlg-examples-appendix}
\end{table*}

\end{document}